\documentclass{article}
\usepackage{arxiv}
\usepackage[utf8]{inputenc}
\usepackage{amssymb,amsthm}
\usepackage{enumitem}
\usepackage{microtype}
\usepackage{idxlayout}
\usepackage{lmodern}
\usepackage{float}
\usepackage{amsthm}
\usepackage{graphicx}
\usepackage{mathtools}
\usepackage{longtable,booktabs}
\usepackage{subcaption}
\usepackage{fullpage}
\usepackage{times}
\usepackage[rightcaption]{sidecap}
\sidecaptionvpos{figure}{t}
\usepackage{fancyhdr,graphicx,amsmath,amssymb}
\usepackage{wrapfig,lipsum,booktabs}
\usepackage[ruled,vlined, linesnumbered]{algorithm2e}
\include{pythonlisting}

\makeatletter
\newcommand{\printfnsymbol}[1]{%
  \textsuperscript{\@fnsymbol{#1}}%
}
\makeatother

\title{reinforcement learning with world model}

\author{
  Jingbin Liu \thanks{equal contribution} \\
  CreateAmind\\
   \And
 Xinyang Gu\printfnsymbol{1} \\
  CreateAmind\\
     \And
 Shuai Liu \\
  CreateAmind\\
}

\begin{document}
\maketitle

\begin{abstract}
Nowadays, model-free reinforcement learning algorithms have achieved remarkable performance on many decision making and control tasks, but high sample complexity and low sample efficiency still hinders the wide use of model-free reinforcement learning algorithms. In this paper, we argue that if we intend to design an intelligent agent that learns fast and transfers well, the agent must be able to reflect key elements of intelligence, like \textbf{Intuition}, \textbf{Memory}, \textbf{Prediction} and \textbf{Curiosity}. We propose an agent framework that integrates off-policy reinforcement learning with world model learning, so as to embody the important features of intelligence in our algorithm design. We adopt the state-of-art model-free reinforcement learning algorithm, Soft Actor-Critic, as the agent intuition, and world model learning through RNN to endow the agent with memory, curiosity and the ability to predict. We show that these ideas can work collaboratively with each other and our agent (RMC) can give new state-of-art results while maintaining sample efficiency and training stability. Moreover, our agent framework can be easily extended from MDP to POMDP problems without performance loss.

\end{abstract}


\section{INTRODUCTION}\label{header-n1}
Model-free deep reinforcement learning algorithms have shown powerful potential in many challenging domains, from tradition board game chess and go to continuous robotic control tasks \cite{that_matters}. 
The combination of reinforcement learning and high-capacity function approximators such as neural networks hold the promise of solving a wide range of tasks, like decision making and robotic control, but when it comes to real-world application, most of these learning methods become impractical. Three major challenges are faced in the current reinforcement learning realm \cite{SAC}.

First of all, model-free learning algorithms suffer from low sample efficiency and high sample complexity, For tasks as simple as gym Box2d games, millions of interaction with the environment are needed. When it comes to complex decision-making problem,
the total interaction steps could easily exceed $10^{10}$, which is inaccessible and unfriendly to most reinforcement learning community researchers. As for projects like Dota2 and Starcraft II, the expenses become formidable. 

Secondly, deep reinforcement learning methods are often extremely brittle with respect to their hyper-parameters. Hyper-parameters tuning is time consuming and computing sources consuming and we need carefully tune the parameters in order to obtain the optimal result. Besides, agent training are likely to stuck at local optimum and many seeds for network weights initialization are indispensable. In some cases, reward signal is not provided or is too sparse to use for the environment we want to solve. We need to handcraft the reward function which brings in more hyper-parameters. Even when the reward signal is relatively dense, good results with stable training are not guaranteed. What is worse, reward design is not general but environment-specific \cite{design}. 

Thirdly, most of the current reinforcement learning algorithms are designed to solve the MDP problems, but problems in the real world seldom follow the MDP assumption. They are often partially observation MDPs (POMDPs), which means that you must make a decision with imperfect information. What makes things worse, It is significantly difficult if not impossible to construct and infer hidden states from the observations given by the environment. The hidden state depends on the entire interaction history of the agent and the environment dynamics which requires substantial domain knowledge of the environment.

In order to mitigate these issues, we propose a new approach to deal with complex tasks with deep reinforcement learning. We name our agent RMC (Recurrent Model with Curiosity). In this paper, we discuss the keys elements of intelligence and attempt to manifest them in a reinforcement learning framework design. From the knowledge of brain mechanisms and psychology, we argue that intuition, curiosity, memory and the ability to predict are the most important if not the entire facets of intelligence and the designed agent can benefit from these ideas with respect to continuous learning and transfer learning.

Like other brain-inspired agent framework, we adopt RNN structure to handle long-term and short-term memory \cite{RDPG}\cite{R2D2}. We investigate and devise a deep-learning approach to learning the representation of states in fully or partially observable environment with minimal prior knowledge of the domain. Then we use the learned hidden states as basis to construct different capabilities of the agent. In particular, we propose a new family of hybrid models that take advantage of both supervised learning and reinforcement learning.  The agent framework can be trained in a joint fashion: The supervised learning component can be a
recurrent neural network (RNN) combined with different heads, providing an effective way of learning the representation of hidden states. The RL component, Soft Actor-Critic (SAC1) \cite{SAC}, learns to act in the environment by maximizing long-term rewards. Furthermore, we design a curiosity bonus from the prediction error, which leads to better representation and exploration.
Extensive experiments on both MDP and POMDP tasks demonstrate the effectiveness and advantages of our proposed agent framework, which outperforms several previous state-of-the-art methods. The key features of our algorithm can be summarized as follow:

\begin{itemize}
\item
  We employ recurrent neural network (GRU) and R2D2, an efficient data structure for hidden states, to learn the
  representation for RL. Since these recurrent models can
  aggregate partial information in the past and capture long-term
  dependencies in the sequence information, their performance is
  expected to be superior compared to the contextual window-based approaches, which are widely used in reinforcement learning models, such as DQN model \cite{DQN}.
\item
  In order to leverage supervised signals in the training
  data, our method combines supervised learning and reinforcement learning and the proposed hybrid agent framework is jointly trained using stochastic gradient descent (SGD): in each iteration, the representation of the hidden states is first
  learned using supervised signals (i.e. next observation and reward)
  in the training data; then, the value function and policy function are updated using SAC1 which takes the learned hidden states as input. By jointly training the world model and acting policy, we can obtain a better representation to capture the environment dynamics.
\item
  Finally, in order to encourage our agent to explore more, we use a curiosity bonus as the decaying internal reward in addition to the external reward. The reward bonus is design according to the prediction error of the learned world model. We expect our agent, guided by the curiosity reward, to avoid local optimum and reach the goal robustly.
\end{itemize}

\section{BACKGROUND}\label{header-n2}

\subsection{POMDP}\label{header-n21}

The key problem of reinforcement learning is to search for an optimal policy that maximizes the cumulative reward 
in Markov Decision Process (MDP) or Partially Observable Markov Decision Process \((\text{POMDP})\). MDP can be seen as a special case of POMDP. POMDP can be defined by the tuple \((\mathcal{S,A,P,R},\Omega,\mathcal{O})\) \cite{planet}, while MDP is defined by \((\mathcal{S,A,P,R})\). The meanings of the notations are as follows:

\begin{itemize}
\item
  \(\mathcal {S}\) represents a set of states
\item
  \(\mathcal {A}\) represents a set of actions, 
\item
  \(\mathcal {P:S\times A\to P(S)}\) stands for the transition function which maps
  state-action to the probability distributions of the next state
  \(\text{P}(s^{\prime}|s,a)\)
\item
  \(\mathcal {R}:\mathcal{S\times A\times S}\to \mathbb{R} \) corresponds to the
  reward function, with \(r_t=r(s_t,a_t,s_{t+1})\)
\item
  \( \Omega\) gives the observations potentially received by the
  agent 
\item
  \(\mathcal{O}\) gives the observation probabilities conditioned on action and next state.
\end{itemize}

Within this framework, the agent acts in the environment according to \(a \in \mathcal{A}\).
the environment changes to a new state following $s^{\prime} \sim \mathcal{P}(\cdot|s, a)$. Next,
an observation \(o \in \mathcal{O}\) and reward $r \sim \mathcal{R}(s,a)$ are received by the agent. The observation may only contain partial information about the underlying state \(s \in \mathcal{S}\). 
Thus we can additionally define an inference model as  \(\text q(s_t|o_{\leq t},a_{<t})=\text{q}(s_t|s_{t-1},a_{t-1},o_t)\)

Although there are many approaches suitable for POMDP process, we focus on using
recurrent neural networks (RNNs) with backpropagation through time
(BPTT) to learn a representation of the state for the POMDP. The Deep Q-Network agent (DQN)\cite{DQN} 
learns to play games from the Atari-57 benchmark by training a convolutional network to represent a value function through Q-learning. The agent takes a frame-stacking of 4 consecutive frames as observations and the training data is continuously collected in a replay buffer. Other algorithms like the A3C,
use an LSTM and the training directly uses the online stream of experience
without using a replay buffer. In paper \cite{RDPG}, DDPG is combined 
with an LSTM by storing sequences in the replay and initializing the
recurrent state to zero during training.







\subsection{ENTROPY-REGULARIZED REINFORCEMENT LEARNING}\label{header-n22}


Within the entropy-regularized framework \cite{entropy_rl}, along with environment reward our agent gets a bonus reward proportional to the entropy of the policy at each time-step as well. This changes the RL problem to:

\begin{equation}
    \pi^* = \arg \underset{\pi}\max \underset{\tau \sim \pi} {\text{E}} \bigg[{ \sum_{t=0}^{\infty} \gamma^t \bigg( r(s_t, a_t, s_{t+1}) + \alpha \text{H}\left(\pi(\cdot|s_t)\right) \bigg)}\bigg]
\end{equation}{}

The temperature parameter $\alpha$ makes trade-off between the importance
of the entropy bonus against the environment's reward.
When $\alpha$ is large, the policy tend to have larger entropy, which means the policy will be more stochastic, on the contrary,
if $\alpha$ become smaller, the policy will become more deterministic.

In the entropy-regularized framework, \(V^{\pi}\) and \(Q^{\pi}\) should be modified to include the entropy term: 
\begin{align}
    V^{\pi}(s) &= \underset{\tau \sim \pi} {\text{E}} \bigg[{ \left. \sum_{t=0}^{\infty} \gamma^t \bigg( r(s_t, a_t, s_{t+1}) + \alpha H\left(\pi(\cdot|s_t)\right) \bigg) \right| s_0 = s}\bigg]  \label{eq:v} \\
    Q^{\pi}(s,a) &= \underset{\tau \sim \pi} {\text{E}}\bigg[{ \left. \sum_{t=0}^{\infty} \gamma^t \bigg( r(s_t, a_t, s_{t+1}) + \alpha  H\left(\pi(\cdot|s_t)\right) \bigg) \right| s_0 = s, a_0 = a}\bigg]
    \label{eq:q}
\end{align}
With equation(\ref{eq:v}, \ref{eq:q}), we can draw the connection between \(V^{\pi}\) and \(Q^{\pi}\). Meanwhile, we have the Bellman equation for \(V^{\pi}\) and \(Q^{\pi}\):
\begin{align}
    V^{\pi}(s) &= \underset{a \sim \pi}{\text{E}}[{Q^{\pi}(s,a)}] \\
    Q^{\pi}(s,a) &= \underset{s' \sim P}{\text E}[{r(s,a,s') + \alpha  H\left(\pi(\cdot|s)\right) + \gamma V^{\pi}(s')}]
\end{align}

\section{METHOD}\label{header-n3}


Inspired by our brain \cite{brain}, we identify four key elements for intelligence, 
i.e.:  \textbf{Intuition}, \textbf{Memory}, \textbf{Prediction} and \textbf{Curiosity}. We use these key elements in our agent design.
The structure of our agent contains three parts: RNN, Model head,
Intuition head, each part has a different role. In the real world, the agent can not observe the full state of the environment. In other words, the Markov property rarely holds in real-world
environments and the tasks are often featured as incomplete and noisy state information because of partial observability.
So we adopt an RNN-based architecture to capture  the past information or experiences that is useful for future decision making. 

We use RNN for inference and memory, which means learning a network to model
\(\text q(s_t|o_{\leq t}, a_{<t})\). In practice we only provide
\(o_t, a_{t-1}\) at time-step \(t\) instead of \(o_{\leq t}, a_{<t}\) so
it is crucial for RNN to store past information into a hidden state
\(s_{t-1}\). The informative hidden state \(s_t\) is also used as
an input of our RL algorithm. It is hard to identify the \(s_t\) as the
"true" state which follows MDP property, but we can push the \(s_t\)
toward the "true" state by jointly training the agent with intuition head and model
head (as discussed in detail in section \ref{header-n32}).

The main function of the model head is to provide a prediction of the future and use
the predicted error as a curiosity bonus. As we all know the ability to understand
and predict the future is fundamental for human. By putting a world model in
our agent, we expect the agent to learn a better representation of hidden state and avoid local optimum.

As for the intuition head, its key function is decision making. In this work, we choose SAC1 as our intuition head. SAC1 is based on the actor-critic framework and uses entropy regularization to encourage exploration. We combine the original algorithm with our model head and curiosity bonus to underlie the intuition for more robust decision making.

\begin{figure}[htbp]
    \centering
    \includegraphics[width=0.6\textwidth]{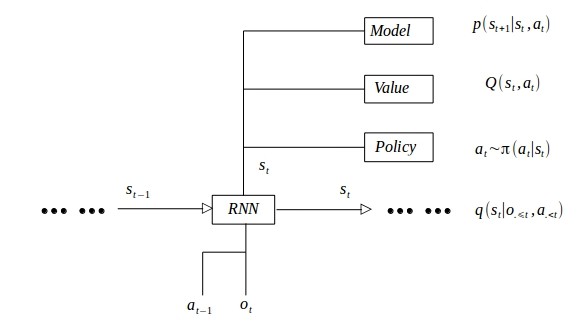}
    \caption{Overall architecture of our RMC agent, which is composed of three main parts: RNN, Model head, Intuition head. 
    The RNN cell infers current state from previous state (\textbf{Memory}) and current observation. 
    Model head is accountable for \textbf{Prediction, Curiosity}. \textbf{Intuition} head has two sub-heads: value head and policy head}
    \label{fig:flow}
\end{figure}{}


\subsection{MEMORY}\label{header-n31}  

\subsubsection{RNN and INFERENCE}
The main function of RNN is providing memory and inference. Due to the POMDP
process, we can't use observation at step \(t\) directly to make
a decision or make a prediction, so we need an inference model to encode
observation, action into a hidden state.
\begin{equation}
    p(s_t|o_{\leq t}, a_{<t}) \Rightarrow p(s_t|s_{t-1},o_t,a_{t-1})
\end{equation}
Since the model is non-linear, we cannot directly compute the state
posteriors that are needed for parameter learning. Instead we can optimize
the function approximator by backing up losses through the RNN cell.
Gradients coming from the policy head are blocked and only gradients
originating from the Q-network head and model head are allowed to
back-propagate into the RNN. We block gradients from the policy head for
training stability, as this avoids positive feedback loops between
\(\pi\) and \(q_i\) with shared representations.
We will discuss the choice of back-propagating value loss and model loss later in section
\ref{header-n34}. In general, training RNN with model and value
function jointly can help with a better representation of the hidden state.

\subsubsection{RECURRENT EXPERIENCE REPLAY}\label{header-n22}

In order to achieve good performance in a partially observed
environment, an RL agent requires a state representation that encodes
information about its state-action trajectory in addition to its current
observation. The most common way to achieve this is using an RNN as part of the
agent's state encoding. So as to train an RNN from replay and enable it to
learn meaningful long-term dependencies, the whole state-action trajectories
need to be stored in the buffer and used for training the network. Recent work \cite{R2D2}
compared four strategies of training an RNN from replayed experience: 
\begin{itemize}
    \item 
      \textbf{Zero start state}: Using a zero start state to initialize the network at the beginning 
        of sampled sequences. 
    \item 
      \textbf{Episode replay}: Replaying whole episode trajectories.
    \item
      \textbf{Stored state}: Storing the recurrent state in the replay and using it to
      initialize the network at training time. 
    \item
      \textbf{Burn-in}: Allow the network a `burn-in period' by using a portion of
      the replay sequence only for unrolling the network and producing a
      start state, and update the network only on the remaining part of the
      sequence. 
    
\end{itemize}{}

The zero start state strategy's appeal lies in its
simplicity and it allows independent decorrelated sampling of
relatively short sequences, which is important for robust optimization
of a neural network. On the other hand, it forces the RNN to learn to
recover meaningful predictions from an atypical initial recurrent state, which may limit its ability to fully rely on its recurrent state and learn to exploit long temporal correlations. 
The second strategy, on the other hand, avoids the problem
of finding a suitable initial state, but creates a number of practical, 
computational, and algorithmic issues due to varying and
potentially environment-dependent sequence length, and higher variance
of network updates because of the highly correlated nature of states in
a trajectory when compared to training on randomly sampled batches of
experience tuples. 
\cite{rdqn} observed little difference
between the two strategies for empirical agent performance on a set of
Atari games, and therefore opted for the simpler zero start state
strategy.

One possible explanation for this is that in some cases, an
RNN tends to converge to a more `typical' state if allowed a certain
number of `burn-in' steps, and so recovers from a bad initial recurrent
state on a sufficiently long sequence. While the zero start state strategy may suffice in the most fully observable
Atari domain, it prevents a recurrent network from learning actual
long-term dependencies in more memory-critical domains.

To fix these issues, Stored state with Burn-in is proposed by \cite{R2D2}.
Empirically, it translates into noticeable performance improvements, 
as the only difference between the pure zero states and the burn-in strategy lies in the fact that the latter unrolls the network over a prefix of states on which the network does not receive updates. 
the beneficial effect of burn-in lies in the fact that it prevents `destructive updates' to the RNN
parameters resulting from highly inaccurate initial outputs on the first
few time steps after a zero state initialization. 
The stored state strategy, on the other hand, proves to be overall much more effective at
mitigating state staleness in terms of the Q-value discrepancy, which
also leads to clearer and more consistent improvements in empirical
performance. 

In all our experiments we use the proposed agent architecture
with replay sequences of length $l_{train} = 15$, with an burn-in prefix of $l_{burn-in} = 10$. 

\subsection{PREDICTIVE MODEL}\label{header-n32}
Humans have a powerful mental model. All human knowledge are essentially predictive physical models. Finding and Learning these models are crucially important to intelligent agents. When it comes to decision making, our intuition can provide actions without planning, but a world model can guide our intuition.  Specifically, in our agent design the role of the world model is to predict the future. Base on the learned hidden state in the model learning, we develop our intuition. 
Meanwhile, the model prediction loss serves as an curiosity to help explore the world. We hypothesize by jointly training on model loss and intuition loss, we can get a better representation.

The transition dynamics can be written as $s' \sim \mathrm{P}(\cdot|s,a)$. Here we use 
an function approximator modeled with a feed-forward neural network 
$s_{t+1}=\hat{f}_\psi(s_t,a_t)$  to capture  the latent dynamic.
We predict the change in state $s_{t+1}-s_t$, given a state $s_t$ and an action $a_t$ as inputs. 
This relieves the neural network from memorizing the input state, especially when the change is small \cite{metrpo}. 
Instead of using the \(L2\) one-step prediction loss as model loss, here we use the $L1$ loss as proposed in \cite{SLBO}, which also outperforms the $L2$ loss in our experimental results.
\begin{equation}
    L_{model}(\psi) = \frac{1}{|\mathcal{D}|}\underset{(s_t,a_t,s_{t+1}\in\mathcal{D})}\sum\|s_{t+1}-\hat{f}_{\psi}(s_t,a_t)\|_2
    \label{eq:model}
\end{equation}


\subsection{CURIOSITY}\label{header-n322}

There are many types of curiosities, but the most fundamental one is the curiosity which is inspired when we can't predict correctly \cite{curiosity}. In our agent design, we use the prediction error as curiosity. In addition to the extrinsic reward, curiosity-driven intrinsic reward is defined as $r^i_t = \hat{s}_t - s_t$.
Let the intrinsic curiosity
reward generated by the agent at time t be \(r^i_t\) and the extrinsic
reward be \(r^e_t\). The policy is trained to maximize the sum of the two kinds of reward:
\begin{equation}
    R(s_t,a_t,s_{t+1})=\beta \cdot r^i_t + r^e_t
\end{equation}{}
We use a parameter \(\beta\) to represent the strength of
intrinsic reward. In practice, We decay the parameter \(\beta\) to zero as the learning process goes on. With the intrinsic reward, the agent is encouraged to explore more in the earlier stage of training and to focus on performance in the later stage of training.

\subsection{INTUITION}\label{header-n33}

The intuition is implemented by applying the SAC1 algorithm \cite{SAC1}. Function approximators is used for both the Q-functions \( Q_{\phi_1}, Q_{\phi_2}\) and the policy \(\pi_{\theta}\).
Instead of running evaluation and improvement to convergence,
we alternate between optimizing both networks with stochastic gradient descent.

\subsubsection{LEARNING Q-FUNCTIONS}
The Q-functions are learned by applying the Bellman backups. Like in TD3 \cite{TD3} and SAC1 \cite{SAC1}, tow target Q networks are used, the backup Q target is defined by the minimum of the two targets.
As for the target network, we obtain it by Polyak averaging the value network parameters in the course of training.
The loss for Q-function can be written:
\begin{equation}
    L_{Q}(\phi_i, {\mathcal D}) = \underset{\substack{s \sim \mathcal{D} \\ \tilde{a} \sim \pi_{\theta}}} {\text{E}}{\Bigg[ \Bigg(Q_{\phi_i}(s,a) - \left(r + (1-d)(\min_{i=1,2} Q_{\phi_i}(s,\tilde{a}) - \alpha \log \pi_{\theta}(\tilde{a}|s)) \right)\Bigg)^2 \Bigg]} \label{eq:qvalue}
\end{equation}{}

The difference between the above soft Q backup and the standard Q backup is the entropy term. The The Q functions are updated in a off-policy fashion. A replay buffer is used for storing \((s, a, s^\prime, r, d)\) tuples. Training data are randomly sampled from the replay buffer.

\subsubsection{LEARNING THE POLICY}
In SAC1 \cite{SAC1}, the policy improvement is conducted based on the learned Q functions. The policy network is represented as an Gaussian distribution. In order to make the policy differentiable, which is crucial for back-propagation, we apply the reparameterization trick.
As suggested in SAC1 \cite{SAC1}, we use a squashed Gaussian
policy, which samples actions according to:
\begin{equation}
    \tilde{a}_{\theta}(s, \xi) = \tanh\left( \mu_{\theta}(s) + \sigma_{\theta}(s) \odot \xi \right), \;\;\;\;\; \xi \sim \mathcal{N}(0, I)
\end{equation}{}
It should be noticed that the variance is local, instead of a global variance used in most other algorithms. The variance  varies with each individual state, which provides more representation capacity. The reparameterization trick allows us to convert the action sampling into Gaussian sampling. The policy loss can be written as:
\begin{equation}
    L_{policy}(\theta) = \underset{\substack{s \sim \mathcal{D} \\ \xi \sim \mathcal{N}}}
    {\text{E}}[{Q_{\phi_1}(s,\tilde{a}_{\theta}(s,\xi)) - \alpha \log \pi_{\theta}(\tilde{a}_{\theta}(s,\xi)|s)}]
    \label{eq:policy}
\end{equation}{}

\subsubsection{LEARNING THE TEMPERATURE}

The temperature $\alpha$ is actually the reward scale of the entropy bonus. Thus it is the ratio of the temperature $\alpha$ and reward scale that determine the extent of exploration. $\alpha$ can be tuned for a specific environment. We can also adapt $\alpha$ during training by targeting a given entropy $\overline{\mathcal{H}}$ for the policy, as suggested in \cite{SAC1}.
The learning of the temperature $\alpha$ is achieved by computing the gradient for \(\alpha\) with the following objective:
\begin{equation}
    L_{temp}(\alpha) = \text{E}_{a \sim \pi_t}[-\alpha \log \pi_t(a_t|s_t) - \alpha \overline{\mathcal{H}}]
    \label{eq:alpha}
\end{equation}{}

The final algorithm is presented in Algorithm \ref{algo:1}. 
The method alternates between collecting experience
from the environment with the current policy and updating the function approximators using the
stochastic gradients from batches sampled from the replay buffer. The agent is trained according the model loss and intuition loss in a concise way.

\begin{algorithm}[t]
\DontPrintSemicolon
\SetAlgoLined
\SetKwInOut{Input}{Input}\SetKwInOut{Output}{Output}
\Input{Initial policy parameters $\theta$ \\
\BlankLine
Transition model parameter $\psi$
\BlankLine
Q-function parameters $\phi_1$, $\phi_2$\\
\BlankLine
Temperature $\alpha$ \\
\BlankLine
Empty replay buffer $\mathcal{D}$
}
\BlankLine
 Set target parameters equal to main parameters $\overline{\theta} \leftarrow {\theta}$, $\overline{\phi}_1 \leftarrow {\phi}_1$, $\overline{\phi}_2 \leftarrow {\phi}_2$ 
 
\BlankLine
\While{not converge}{    
    
    \BlankLine
    
    \For{each environment step}{
        \BlankLine
        $a_t \sim \pi_{\theta}(a_t|s_t)$ \tcp*{Sample action from the policy}
        \BlankLine
        $s_{t+1} \sim p(s_{t+1}|s_t, a_t)$\tcp*{Sample transition from the environment}
        \BlankLine
        $\mathcal{D} \leftarrow \cup \{(o_t, s_t, a_t, r(s_t, a_t), o_{t+1}, s_{t+1}\}$ \tcp*{Store the transition in the replay pool}}
        
    \For{each gradient step}{
    $\{(s, a, r, s^\prime, d)\}^B_{i=1} \sim \mathcal{D}$ \tcp*{Randomly sample a batch of transitions} 
    \BlankLine
    Compute model loss $L_{model}(\psi)$ from equation(\ref{eq:model})
    \BlankLine
    $\psi \leftarrow \psi - \lambda_{\psi} \bigtriangledown_{\psi} L_{model}(\psi)$ \tcp*{Update model parameter} 
    
    \BlankLine
    Compute Q-value loss $L_{Q}(\phi)$ from equation(\ref{eq:qvalue})
    \BlankLine
    $\phi \leftarrow \phi - \lambda_{\phi} \bigtriangledown_{\phi} L_{Q}(\phi)$ \tcp*{Update Q-value parameter} 
    
    \BlankLine
    Compute policy loss $L_{policy}(\theta)$ from equation(\ref{eq:policy})
    \BlankLine
    $\theta \leftarrow \theta - \lambda_{\theta} \bigtriangledown_{\theta} L_{policy}(\theta)$ \tcp*{Update policy parameter}

    \BlankLine
    Compute temperature loss $L_{temp}(\alpha)$ from equation(\ref{eq:alpha})
    \BlankLine
    $\alpha \leftarrow \alpha - \lambda_{\alpha} \bigtriangledown_{\alpha} L_{temp}(\alpha)$ \tcp*{Update temperature  parameter} 
    \BlankLine
    $\overline{\phi}_i \leftarrow \rho {\overline{\phi}}_i + (1-\rho) {\phi}_i, \ \ \text{for} \ i \in \{1,2\}$ \tcp*{Update target network weights}

    }

}
\caption{RMC AGENT}
\label{algo:1}
\end{algorithm}

\section{EXPERIMENT}\label{header-n4}

In order to test our agent, We designed our experiments to answer the following
questions:
\begin{enumerate}
    \item 
    Can RMC solve challenging continuous control
    problems? How does it perform with regard to the final performance and sample complexity, comparing with other state-of-the-art methods?
    \item
    Can RMC handle POMDP environments? How well does it deal with the absence of information and generalize?
    \item
    We optimize RNN on model loss and intuition loss jointly. Does the model learning really help us improve the agent's performance?
    \item
    Can we benefit from the decaying curiosity bonus, which is based on the model prediction loss?
    
\end{enumerate}{}
To answer (1), we compare the performance of our agent with other state-of-the-art methods in section \ref{header-n41}. 
To answer (2), we purpose a modified mujoco environment, the flicker mujoco, which is a POMDP environment. We will discuss the details of the environment and the experiment setting in section \ref{header-n42}. 
With regard to (3) and (4), we address the ablation study on our algorithm in section \ref{header-n43}, 
testing the influence of different training schemes and network designs on the agent's final performance.


\subsection{MUJOCO}\label{header-n41}
The goal of this experimental evaluation is to understand how the sample
complexity and stability of our method compares with previous off-policy
and on-policy deep reinforcement learning algorithms. We compare our
method to previous techniques on a range of challenging continuous control
tasks from the OpenAI gym benchmark suite. Although easier tasks can
be solved by a wide range of different algorithms, the more complex
benchmarks, such as the 21-dimensional Humanoid, are exceptionally difficult 
to solve with off-policy algorithms. For easier tasks, most of reinforcement algorithms can achieve good results by tuning hyper-parameters. While for the hard tasks, already narrow basins of effective hyper-parameters become prohibitively small for hyper-parameters sensitive algorithms. The hard tasks can be effective algorithm test beds \cite{QProp}.

We compare our method with deep deterministic policy gradient (DDPG) \cite{ddpg}, an algorithm that is regarded as an expansion to continuous action space of DQN algorithm; policy optimization (PPO) \cite{ppo}, a stable and
effective on-policy policy gradient algorithm; and soft actor-critic(SAC1) \cite{SAC1}, 
a recent off-policy reinforcement learning algorithm with entropy regularization.
We additionally compare our method with  twin delayed deep deterministic
policy gradient algorithm (TD3) \cite{TD3}.

We conduct the robotic locomotion experiments by using the MuJoCo simulator \cite{mujoco}. The states of the
robots are their generalized positions and velocities, and the
controls are joint torques. High dimensionality and non-smooth dynamics due to contacts make these tasks very challenging. 
To allow for a reproducible and fair comparison, 
we evaluate all the algorithm with a similar network structure. 
For the off-policy algorithm, we use a two-layer feed-forward neural network 
of 400 and 300 hidden nodes respectively, with rectified linear units (ReLU)
between each layer for both the actor and critic. 
For on-policy algorithm, we use the parameters which is shown in previous work \cite{that_matters} as a comparison of our agent. 
Both network parameters are updated using Adam\cite{adam} with a learning rate of $10^{-4}$. No modifications are made to the environment and reward.

\begin{figure}
    \centering
    \begin{subfigure}[b]{0.3\textwidth}
        \includegraphics[width=\textwidth]{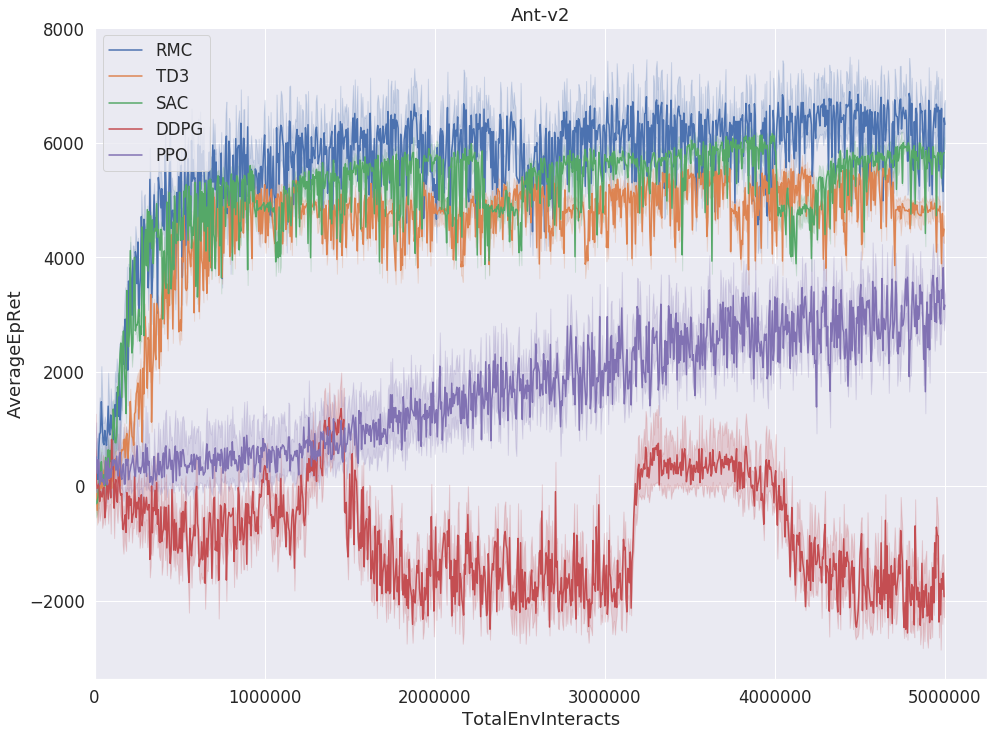}
        \caption{Ant-v2(MDP)}
        \label{fig:antmdp}
    \end{subfigure}
    \begin{subfigure}[b]{0.3\textwidth}
        \includegraphics[width=\textwidth]{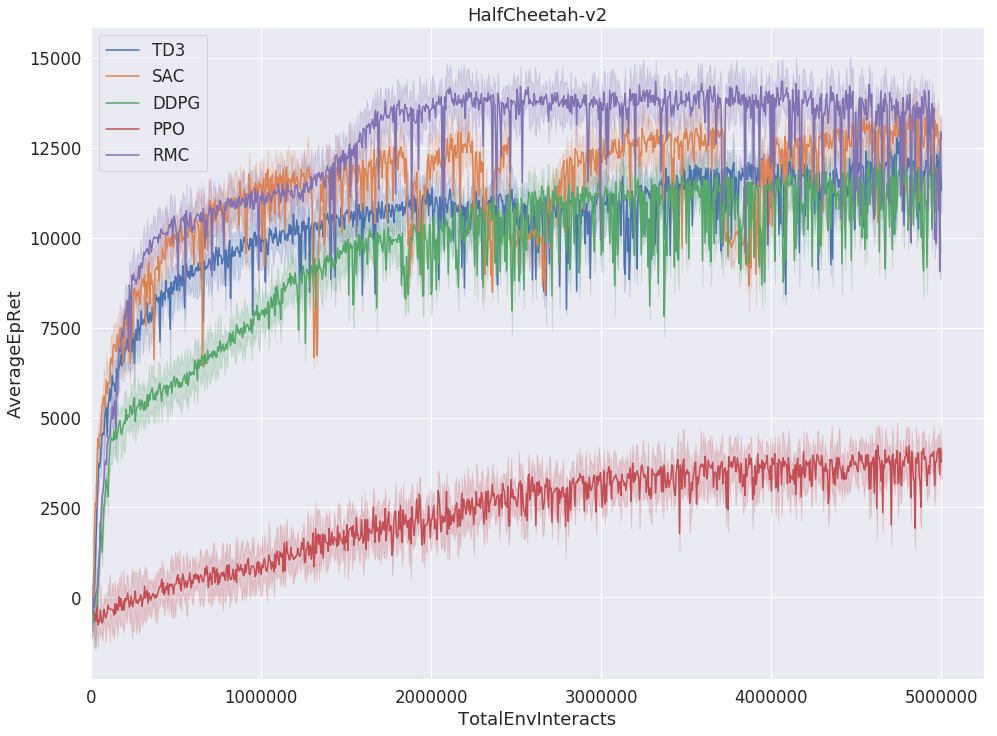}
        \caption{HalfCheetah-v2(MDP)}
        \label{fig:hcmdp}
    \end{subfigure}
    \begin{subfigure}[b]{0.3\textwidth}
        \includegraphics[width=\textwidth]{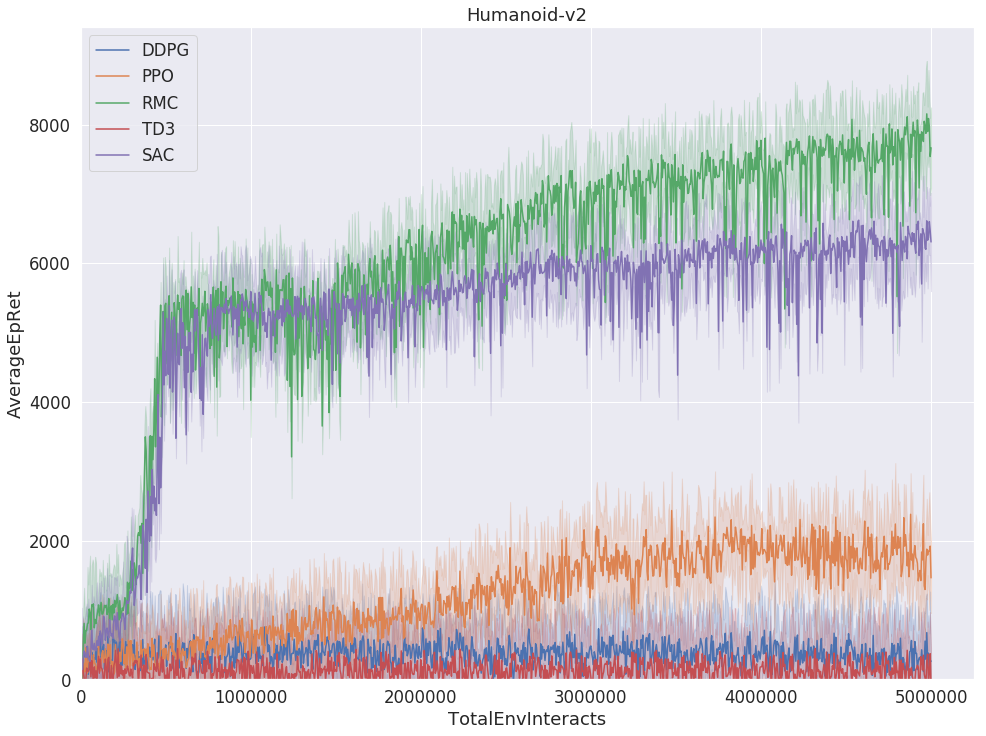}
        \caption{Humanoid-v2(MDP)}
        \label{fig:hummdp}
    \end{subfigure} 
    \medskip
    \begin{subfigure}[b]{0.3\textwidth}
        \includegraphics[width=\textwidth]{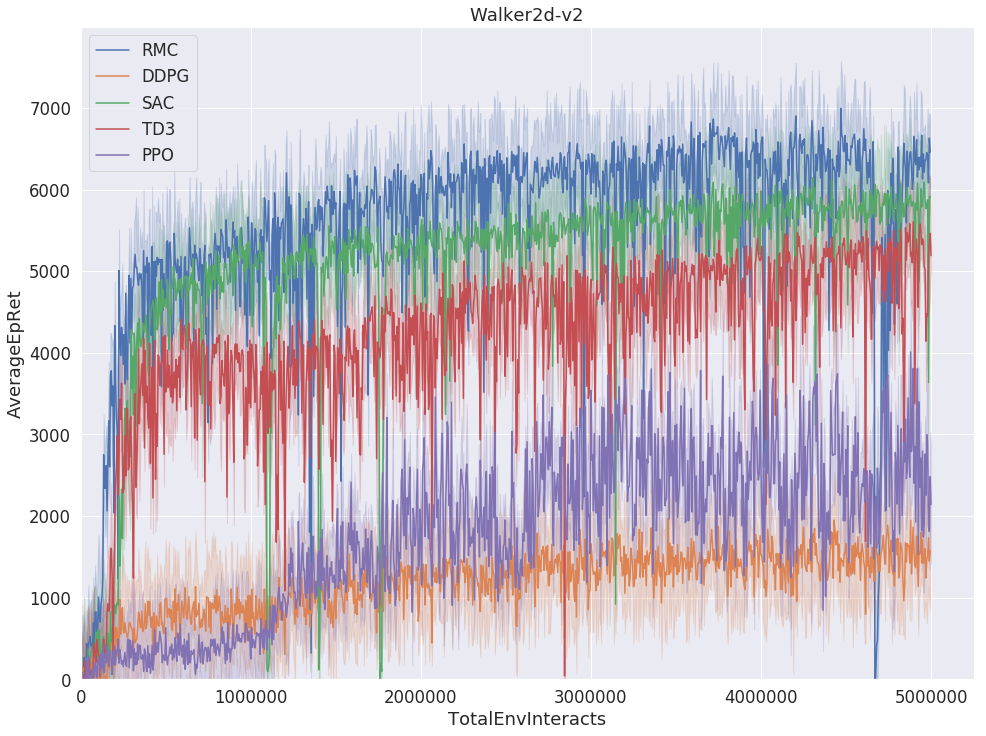}
        \caption{Walker2d-v2(MDP)}
        \label{fig:antpo}
    \end{subfigure}
    \begin{subfigure}[b]{0.3\textwidth}
        \includegraphics[width=\textwidth]{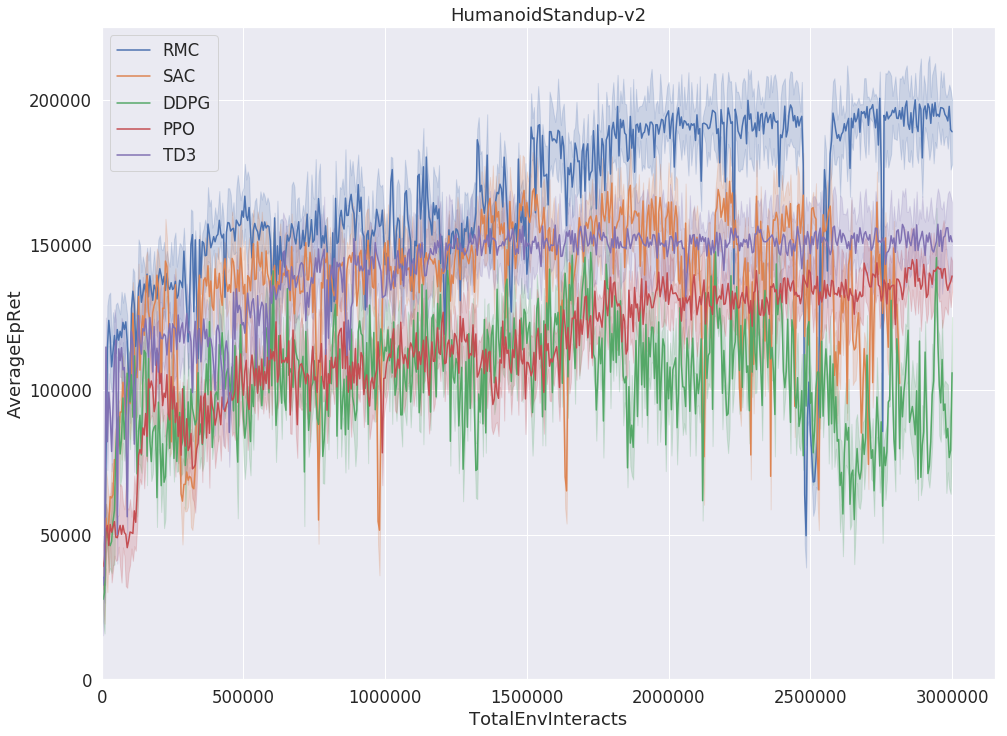}
        \caption{HumanoidStandup-v2(MDP)}
        \label{fig:hcpo}
    \end{subfigure}
    \begin{subfigure}[b]{0.3\textwidth}
        \includegraphics[width=\textwidth]{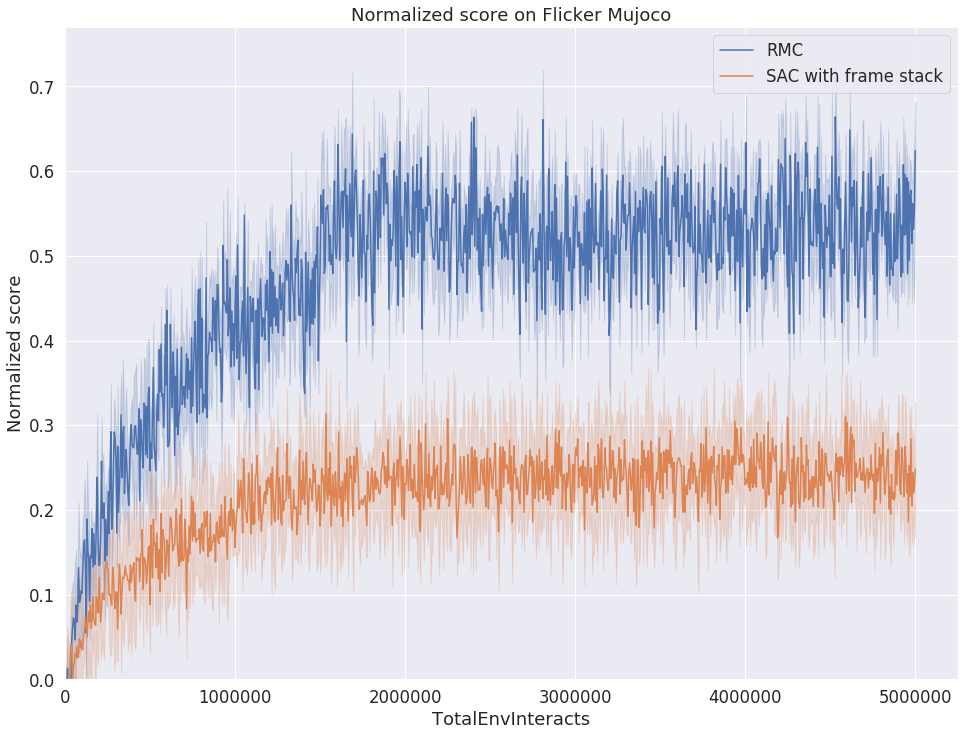}
        \caption{Flicker MuJoCo(POMDP)}
        \label{fig:humpo}
    \end{subfigure}
    
    \caption{ (a) to (e) are the training curves on continuous control benchmarks. RMC agent performs well
    consistently across all tasks and outperforms both on-policy and off-policy methods in the challenging MuJoCo tasks.
    (f) shows the noramlized score for training on Flicker MuJoCo with $p=0.5$. }
    \label{fig:mujoco}
\end{figure}

In Fig. \ref{fig:mujoco}, we compare our method with different other methods in five individual runs initialized with different random seeds. RMC significantly outperforms the baselines, indicating substantially better efficiency and stability.
It shows that the model training and the intrinsic curiosity return can actually promote the policy learning of the agent.

\subsection{POMDP MUJOCO}\label{header-n42}
In order to test our agent's ability to deal with imperfect information cases, we establish a new environment
called \emph{Flickering MuJoCo}, which is achieved by changing classic MuJoCo benchmark to a POMDP environment.
At each time-step, the screen is either fully revealed or fully obscured with a probability of \(p=0.5\). Obscured frames convert MuJoCo into a POMDP MuJoCo in a probabilistical manner. 

Many previous works deal with POMDP by using a long history of observations, while in our RMC agent we use a recurrent network trained with a single observation at each timestep. As shown in Fig. \ref{fig:humpo}, our agent outperforms standard SAC1 combine with frame stack. Our agent performs well at this task even when given only one input frame per time-step. RMC can successfully integrates noisy information through time to detect critical events.

\begin{figure}
    \centering
    \begin{subfigure}[b]{0.3\textwidth}
        \includegraphics[width=\textwidth]{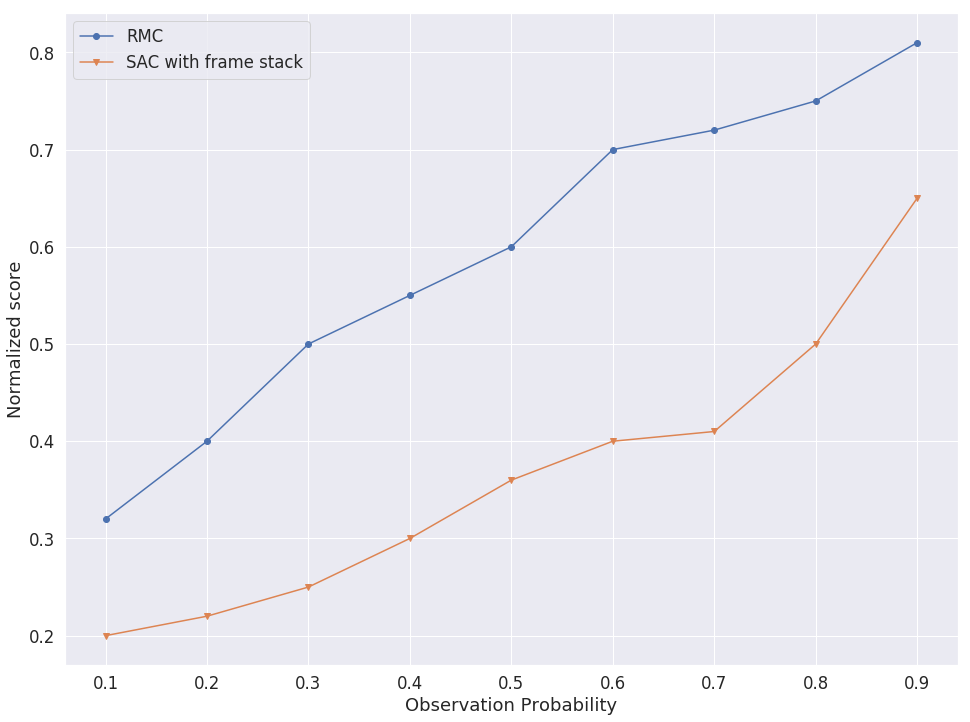}
        \caption{from MDP to POMDP}
        \label{fig:mtp} 
    \end{subfigure}
        \begin{subfigure}[b]{0.3\textwidth}
        \includegraphics[width=\textwidth]{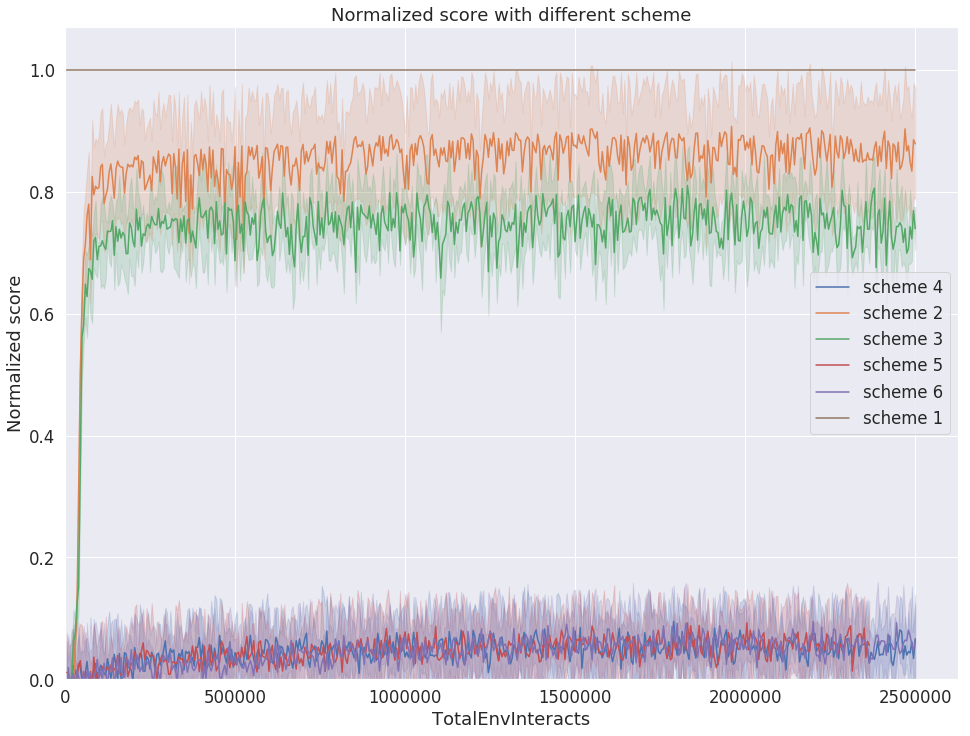}
        \caption{6 training schemes}
        \label{fig:uc}   
    \end{subfigure}
    \begin{subfigure}[b]{0.3\textwidth}
        \includegraphics[width=\textwidth]{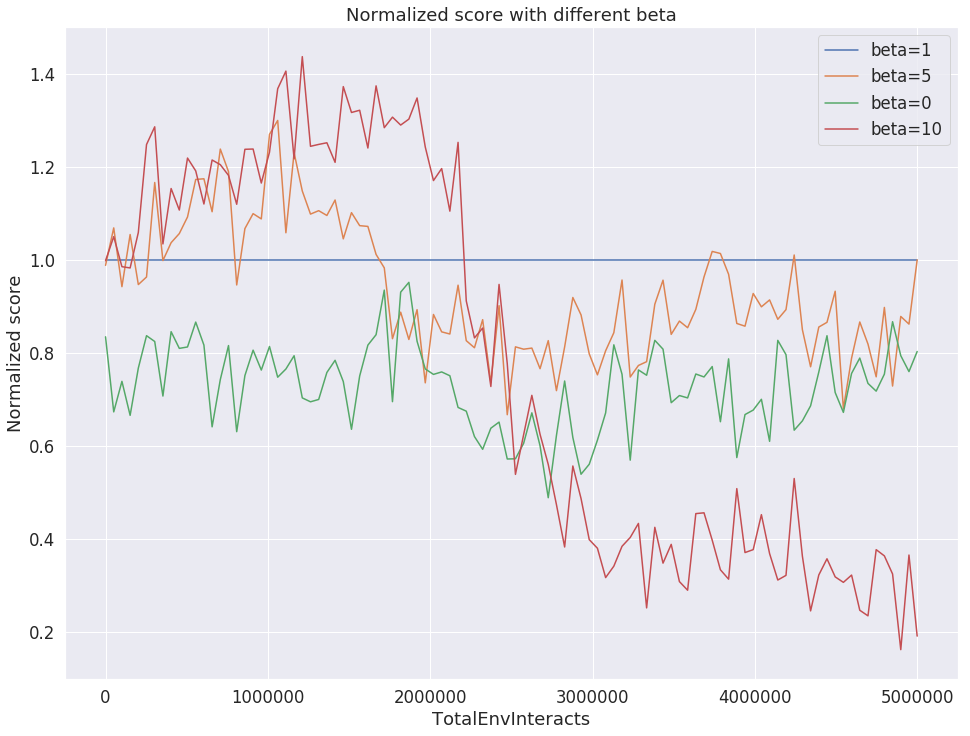}
        \caption{curiosity}
        \label{fig:design} 
    \end{subfigure}
    \caption{(a) Agents are first trained on MDP environments and then
            evaluated on POMDP environments. RMC's performance loss over obscured observation is less than SAC1. Each data
            point stands for the normalized score over different flickering probability, evaluated on environments shown in Fig. \ref{fig:mujoco}. 
            (b) We test 6 different training schemes on HalfCheetah-v2. It shows that updating RNN on model loss, value loss and not on policy loss gets the best result.
            (c) The impact of different scales of curiosity reward.}
            \label{fig:generalization}
    
\end{figure}

It is interesting to ask a question: Can or to what extent can an agent which is trained on a standard MDP generalize to a POMDP at evaluation time? 
To answer this question, we evaluate the RMC agent and SAC1 trained on standard MuJoCo over the flickering MuJoCo. 
Fig. \ref{fig:mtp} shows our agent preserves more of its previous performance than SAC1 across all levels of flickering. We conclude that RMC is more robust against loss of information.

\subsection{ABLATION STUDY}\label{header-n43}
\subsubsection{THE IMPACT OF DIFFERENT TRAINING SCHEMES}\label{header-n431}

There are three major heads for RMC: the model head, the value head and the policy head. In the learning procedure, it is important to determine which of these head's loss can be back-propagated into RNN. In order to address this problem, we study the impact of different training schemes. We summarize six valid update schemes in Table \ref{table:scheme}.
Theoretically, the model loss alone is sufficient for the representation learning, we can block the policy and value loss from back-propagating into RNN. In practice, the collection of training data is controlled by the intuition, the value or policy losses may be beneficial for stabilizing the training process.

Fig. \ref{fig:uc} shows the learning performances of different update
schemes. For schemes 4, 5 and 6, the policy loss are back-propagated into RNN. We can see from the results that the policy loss can disrupt the learning process greatly. Their policies become nearly random and fails to exploit the reward
signal, resulting in substantial degradation of performance. One possible explanation is that the policy loss is too strong to be used in the model learning.
For schemes 2 and 3, we can see that with the value loss alone or the model loss alone, the performance is enhanced. It is straight forward to infer that scheme 1 can have the best performance with value loss and model loss back-propagating into RNN jointly. For scheme 1, gradient calculated from the policy loss is blocked and only gradients
originated from the value head and model head are allowed to
back-propagate into the RNN. The model head makes sure the representation learning of a state is good enough to predict its next state. At the same time, the value head can inject the information that is needed to learn a good policy to the same representation. Just like our brain has two different kinds of thinking pattern, the so-called intuition thinking and
reasoning thinking, our agent can take advantage of joint optimization of intuition behavior and predictive model, by
learning a representation of state that supports for a intuitive and predictive agent.

In a word,  it is important to assign different tasks to different heads reasonably. The value and model bear the burden of representation learning, while the policy head is only responsible for learning the intuition. This combination can produce best performance in our experimental setting.

\begin{wraptable}{r}{0.4 \textwidth}
\begin{longtable}[c]{@{}clll@{}}
\caption{All valid update schemes} 
\label{table:scheme} \\
\toprule
scheme & Model & Value & Policy\tabularnewline
\midrule
\endhead
1 & \emph{True} & \emph{True} & \emph{False}\tabularnewline
2 & \emph{False} & \emph{True} & \emph{False}\tabularnewline
3 & \emph{True} & \emph{False} & \emph{False}\tabularnewline
4 & \emph{True} & \emph{True} & \emph{True}\tabularnewline
5 & \emph{True} & \emph{False} & \emph{True}\tabularnewline
6 & \emph{False} & \emph{True} & \emph{True}\tabularnewline
\bottomrule
\end{longtable}
\end{wraptable} 

\subsubsection{THE IMPACT OF CURIOSITY}\label{header-n432}
As discussed in section \ref{header-n3}, the model head can provide a 
prediction of the future step and in the meanwhile provide a curiosity bonus for policy learning. We've already analyzed the influence of different model update schemes and shown that joint training can improve performance. In this section, we discuss the impact of curiosity on the agent training.

We choose update scheme 1 to conduct the experiments on curiosity.
As illustrated in Fig \ref{fig:design}, when we set $\beta$ to zero, 
the model can't explore well so both the sample efficiency and final score
is below the benchmark. If we use a huge scale of $\beta$, the intrinsic reward dominates the policy signal and prevents the policy from utilizing external reward for learning.

Theoretically, as the learning process goes on, the model loss will become smaller and smaller until zero so that $\beta$ won't have much impact in the end  and all the different $\beta$ scales will lead to a similar final performance. 
But in practice, due to insufficient training and stochasticity of the environment, the model loss can never become zero. In order to achieve fast and stable learning, we decay the $\beta$ scale from large to small, encouraging our agent to explore more at the beginning and exploit more in the end.

\section{CONCLUSION}\label{header-n5}

In this paper, we argue that it is important for an intelligent agent design to embody the key elements reflected in human intelligence like intuition, memory, prediction and curiosity. We show that these ideas can work collaboratively with each other and our agent (RMC) can give new state-of-art results while maintaining sample efficiency and training stability. The representation learning of the environment and its dynamic is at the core of training an intelligent agent. Joint training of model-free intuition and model-based prediction can improve the representation learning of the task, which underlies and promotes the policy learning.
Empirically results show that our agent framework can be easily extended from MDP to POMDP problems without performance loss, which sheds light on real-world applications. Further study includes incorporating stochastic transition function like planet \cite{planet} and world model \cite{world_model} into our framework and theoretic analysis of the algorithms like our method that exploits model-free and model-based algorithms at the same time.

\bibliographystyle{apalike}
\bibliography{ref.bib}

\end{document}